\documentclass[letterpaper, 10 pt, conference]{ieeeconf}
\IEEEoverridecommandlockouts

\usepackage[english]{babel}

\usepackage{amsmath}
\usepackage{graphicx}
\usepackage[colorlinks=true, allcolors=blue]{hyperref}
\usepackage{comment}
\usepackage[dvipsnames]{xcolor}

\usepackage[font=small]{caption}
\usepackage{cuted}

\title{Enhancing Robustness in Language-Driven Robotics: \\
A Modular Approach to Failure Reduction}% through Policy Alignment}
\author{Émiland Garrabé$^\ast$, Pierre Teixeira$^\dagger$, Mahdi Khoramshahi$^\ast$, Stéphane Doncieux$^\ast$
\thanks{$^\ast$ ISIR, Sorbonne Université, Paris; $^\dagger$ Work done at ISIR. Corresponding email: {\tt\small garrabe@isir.upmc.fr}}
}

\begin{document}
\maketitle
\thispagestyle{empty}
\pagestyle{empty}

%\begin{figure*}[h!]
\begin{strip}
    \centering
    \includegraphics[width=0.9\textwidth]{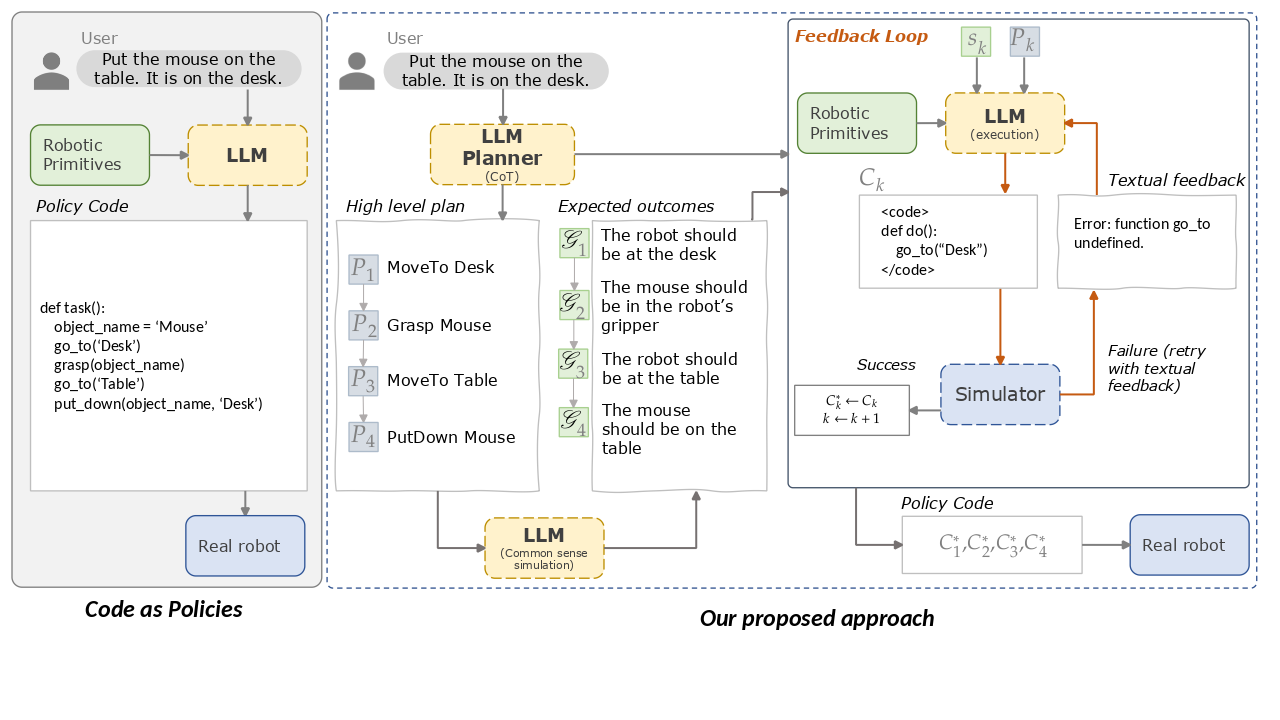}
    \captionof{figure}{\textbf{Overview of the proposed approach:} When receiving instructions from a user, a plan and the expected outcomes (EOs) of each step are inferred. The robot then tackles the task iteratively, using motion primitives and receiving feedback from a simulation, before execution in the environment. See Section \ref{sec:formalism} for notations and formalism.}
    \label{fig:top_fig}
\end{strip}
%\end{figure*}

\begin{abstract}
Recent advances in large language models (LLMs) have led to significant progress in robotics, enabling embodied agents to understand and execute open-ended tasks. 
However, existing LLM-based approaches face limitations in grounding their outputs within the physical environment and aligning with the capabilities of the robot. 
While fine-tuning is an attractive approach to addressing these issues, the required data can be expensive to collect, especially when using very large language models. 
Smaller language models, while more computationally efficient, are less robust in task planning and execution, leading to a difficult trade-off between performance and tractability. 
In this paper, we present a novel, modular architecture designed to enhance the robustness of locally-executable LLMs in the context of robotics by addressing these grounding and alignment issues. We formalize the task planning problem within a goal-conditioned POMDP framework, identify key failure modes in LLM-driven planning, and propose targeted design principles to mitigate these issues. 
Our architecture introduces an ``expected outcomes'' module to prevent mischaracterization of subgoals and a feedback mechanism to enable real-time error recovery. 
Experimental results, both in simulation and on physical robots, demonstrate that our approach leads to significant improvements in success rates for pick-and-place and manipulation tasks, surpassing baselines using larger models. Through hardware experiments, we also demonstrate how our architecture can be run efficiently and locally. 
This work highlights the potential of smaller, locally-executable LLMs in robotics and provides a scalable, efficient solution for robust task execution and data collection.\footnote{Code available at \url{https://tinyurl.com/4h2ysybd}.}
\end{abstract}

\section{Introduction}
Due to their natural language understanding and code writing abilities, large language models (LLMs) are an attractive tool for robots evolving in open environments. This is especially true in service robotics, industrial automation, and human-robot collaboration, where robots can be directed through open-ended language instructions, making them more adaptable and user-friendly. However, LLM-driven robotic systems still face two critical challenges: \textbf{grounding language instructions} in the robot’s physical environment and \textbf{aligning these instructions} with the robot’s actual capabilities. 
Many current methods in robotics rely on very large language models, such as GPT-3.5 ($175$ billion parameters) \cite{you2023robot} or GPT-4o (more than $1.7$ trillion parameters), to achieve reliable performance when executing complex tasks. 
While these models offer high accuracy, they come with significant drawbacks: their substantial computing power and energy requirements lead to high environmental and economic costs, dependence on third-party services for uptime, and potential privacy concerns due to data sharing. 
Despite these issues, comparatively little attention has been given to smaller language models, like LLaMA3.1 (8 billion parameters), which can be executed locally on standard hardware. 
These smaller models, though less powerful, still offer many of the key advantages of large language models, such as the ability to generate code and reason with natural language. 
However, their limited performance exacerbates the grounding and alignment problems. While fine-tuning is often proposed to address such challenges, the design of scalable, zero-shot-compatible data collection architectures remains a challenge.\\
In this paper, we address this challenge by improving the robustness and applicability of smaller language models ($<10$ billion parameters) for efficient and reliable robotic task execution.
Specifically, our contributions are: \textbf{(1)} We formalize the problem of LLM-driven robotic task planning within a goal-conditioned POMDP framework and identify two failure modes: \textbf{subgoal mischaracterization} and \textbf{missing subgoals}. \textbf{(2)} We propose design principles aimed at improving the robustness of the LLM-based planning and execution pipeline. Specifically, we introduce an \textbf{“expected outcomes”} module to prevent errors by inferring detailed descriptions of each plan step and a \textbf{feedback mechanism} to recover from execution failures. \textbf{(3)} Through a series of experiments, both in simulation and on physical robots, we demonstrate that our architecture significantly increases task success rates, even \textbf{outperforming systems using larger language models} while \textbf{incurring significantly lower computational costs}. Our pipeline is also capable of handling instructions given by non-expert users, and, by splitting task execution in several plan steps, is more explainable and can recover from execution failures. We also \textbf{(4)} test one of the dense, distilled versions of Deepseek (Deepseek-r1:7b), exploring its applicability to robotics use cases.

\section{Related works}
\subsection{Foundation models for robotics and planning}
Foundation models are receiving growing attention in planning due to their compatibility with natural language and open vocabularies. Some authors \cite{llmpp, xie2023translating} propose to use LLMs to formalize planning problems and solve them using optimal solvers, and plans inferred by LLMs can be used as a heuristic for a planner\cite{pddlllm}. LLMs can also be used to generate behavior trees \cite{LLMbt} or interact with scene graphs \cite{sayplan} when solving tasks.\\
In the context of robotics, the common sense of LLMs can manifest as an ability to handle ambiguous instructions \cite{cap} or predict user behavior \cite{human_aware}. LLMs can be given access to sets of pre-defined motion primitives \cite{cap, progprompt}, and this in turn can be used to generate datasets to train multitask policies \cite{sup_ddown}. To increase performance, authors propose to decouple the planning and code writing stages \cite{planning_then_code}. LLM-inferred plans can also be used as a basis for reward function writing \cite{robohorizon}. An increasingly popular trend is that of using vision-language models (VLMs) to augment the pipeline with a degree of spatial awareness. For example in COPA \cite{copa}, VLMs are used to detect task-based spatial constraints and this is used for motion planning. VLMs can also be used to propose tasks grounded in the robot's environment to collect task data \cite{autort}.

\subsection{VLA models}
Vision-language-action (VLA) models are a rapidly emerging paradigm in robotics. The key idea of VLAs is to directly predict actions from the task statement and robot camera(s) output\cite{octo}. While early VLA models were trained from scratch, using language models as a backbone has been gaining traction, with the goal of leveraging such models' general knowledge \cite{openvla}, or even retaining multimodal understanding abilities \cite{chatvla}. VLAs have shown impressive results on challenging manipulation tasks, but generalization remains a challenge and they often rely on fine-tuning\cite{pi_0}, requiring large datasets.

\subsection{Grounding and alignment in robotics}
In robotics, the disconnect between the LLM and the environment is an issue. To tackle this, SayCan \cite{brohan2023can} uses a value function score for each predicted action based on its feasibility and likelihood of success. SayPlan \cite{rana2023sayplan} and SayNav \cite{rajvanshi2024saynav} address the grounding problem with iterative re-planning, exploiting a 3D scene graph. Another approach to grounding is to leverage multi-modalities in VLMs and VLAs (\cite{tong2024oval}, see also above).\\
An increasingly popular technique for grounding language models is the implementation of feedback mechanisms. For example, VLM planners can detect and recover from failure \cite{closed_loop}, and human feedback, which can be coupled with uncertainty detection mechanisms \cite{ren2023robots,mullen2024towards} has also been identified as a promising way of grounding LLMs \cite{inner_monologue}. For manipulation task, LLMs have also been used to infer functions measuring the success of uncertain motion primitives \cite{sup_ddown}. Training models \cite{RoboGPT,glam} with domain-specific data is also gaining traction, but collecting enough high-quality data remains a challenge. \\
Studies similar to ours include MALMM \cite{malmm}, where a multi-agent architecture plans and executes a task. However, the architecture doesn't consider explicit expected outcome characterization. Further, it relies on the LLM agents' ability to properly parametrize geometric motion primitives, which heavily relies on the high performance of GPT-4o, leading to scalability issues. Another work \cite{hicrisp} envisions a two-level error management pipeline. However, at the low level, error correction mostly relies on automated fallbacks, while we propose using the execution LLM to handle errors, and using a preprocessing step to prevent them. We finally note that, by design, our pipeline is compatible with high-level plan adaptation techniques.\\

Overall in robotics, foundation models still suffer from issues with alignment and grounding, and most current methods rely on the impressive capabilities of very large models. In this work, we show that feedback mechanisms and explicit subgoal characterization can yield significant increases in performance when using smaller LLMs, and lead to a more explainable, scalable and less computationally demanding system.

\section{Method}\label{sec:formalism}
In this section, we formalize the problem of language-driven task planning and execution. Based on this formalism, we provide novel design principles that allow an LLM-based agent to robustly plan and execute a task.

\subsection{Plan execution as options in a GC-POMDP}
\begin{figure}
    \centering
    \includegraphics[width=0.55\linewidth]{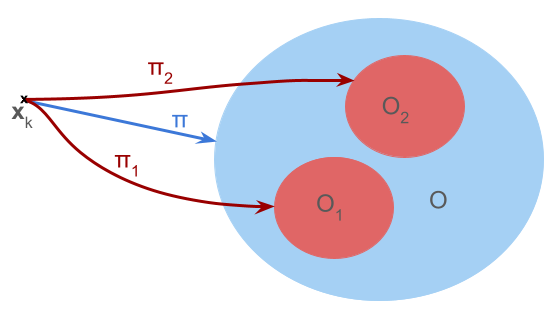}
    \caption{Options $\pi_1$ and $\pi_2$ (red) being parameterized versions of option $\pi$ (blue). The outcome sets of $\pi_1$ and $\pi_2$ are subsets of the outcome set of $\pi$. For example, $\pi$ might be "grasp mug", with $\pi_1$ being "grasp mug handle" and $\pi_2$ "grasp mug body". $O$ denotes the outcome set of $\pi$, while $O_1$ and $O_2$, respectively, denote the outcome sets of $\pi_1$ and $\pi_2$.}
    \label{fig:outcome_subsets}
\end{figure}

We cast the problem of solving a task, expressed in language, as a Goal-Conditioned POMDP (GC-POMDP). A GC-POMDP\cite{gcmdp} is a tuple $(\mathcal{O}, \mathcal{S}, \mathcal{A}, \mathcal{G}, \mathcal{P}, \mathcal{E})$ where $\mathcal{O}$ is the observation space, $\mathcal{S}$ the state space, $\mathcal{A}$ the action space, $\mathcal{G}$ the goal (in our case, the task expressed in language), $\mathcal{P}$ is the transition probability function and $\mathcal{E}$ is the observation probability function. For simplicity, we assume that the reward is $1$ at the end of the episode if the task is complete. We note $\mathcal{G}_s$ the subset of states where the goal is fulfilled.\\
\noindent \textbf{Open-vocabulary LLM planning:} Planning is the problem of finding a sequence of plan steps $\{P_i\}_{i\in 1:n}$, or, equivalently, subgoals $\{\mathcal{G}_i\}_{i\in 1:n}$ such that, when all the $\mathcal{G}_i$'s have been reached in sequence, the task $\mathcal{G}$ is achieved.\\% For each $i$, $\mathcal{G}_{s,i}$ is the set of states where the subgoal is fulfilled.\\
\noindent \textbf{Plan execution:} We consider a set of motion primitives accessible to the robot, modelled as \textit{options}. As per \cite{sutton_options}, an option is a triplet $\langle \mathcal{I}, \pi, \mathcal{B} \rangle$, where $\mathcal{I}$ is the initialization set, the subset of the state space from which the option's policy is defined; $\pi$ is the policy that is followed while the option is active, which we call the intra-option policy and $\mathcal{B}$ is a termination condition. 
We also define $\mathcal{O}$, the set of possible outcomes of the option. This set includes both the successful outcomes $\mathcal{O}_s$ of the option (say for a grasping option, the object is in the robot's gripper) and the failures $\mathcal{O}_f$ (the robot drops the object). 
We consider that the state at the end of the option is uniformly sampled from the union of the two sets. This implies that failures can randomly occur. The designer of the robotic skills doesn't specify their exact outcomes aside from ensuring that they fulfil their purpose, meaning that parametrized options can be designed, where the outcome set is a subset of a more general option's outcome set (see Figure \ref{fig:outcome_subsets}). For example, a grasping option can be parameterized to specify which part of the object is grasped.

Solving the task implies finding a sequence of options such that: (i) the options fulfil each subgoal $\mathcal{G}_i$ sequentially; (ii) the successful outcome set of each option is in the initialization set of the following option: $\mathcal{O}_{s,i} \subset \mathcal{I}_{i+1}$ (where $i$ denotes the index of the option) and (iii) option failure (which can occur randomly) is recovered from.\\
In the following, we will say that an option \textit{is afforded} by a state when this state is in the option's initialization state. While in robotics, affordances (see e.g. \cite{affordance_review, deep_affordance_RL}) are mostly known in the context of manipulation tasks, the concept can be used to describe the possibilities an agent has in a given environment (in our formalism, the options that can be executed from a state).\\
The module responsible for executing the plan steps using the options is the \textbf{execution module}. In this work, the execution module outputs code snippets $C$ in which the options are called through python functions.

\subsection{LLM planning failures}
\begin{figure}
    \centering
    \vspace{0.1cm}
    \includegraphics[width=\columnwidth]{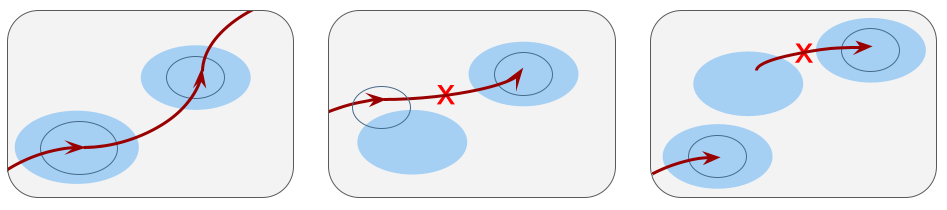}
    \caption{Common errors for LLM plan execution. The GTSGs are plain blue ellipses while the plan subgoals are represented by dark blue contours. The options are in red (we omit the initialization and outcome sets of the options for legibility). Left: the plan subgoals and GTSGs are aligned. Center: Mischaracterized subgoal: while the state achieves the plan subgoal, it does not fulfil the GTSG and does not afford the next primitive. Right: Missing subgoal in the plan: the plan step requires more than one option to be carried out, leading to execution errors.}
    \label{fig:errors}
\end{figure}

In this section, we assume the existence of a set of ground-truth subgoals (or GTSGs) $\{\mathcal{G}_{g,i}\}_{i\in 1:m}$ such that the transition between each pair of successive subgoals $\{\mathcal{G}_{g,i}, \mathcal{G}_{g,i+1}\}$ can be achieved with one motion primitive and the plan obtained by following the GTSGs is valid. After surveying LLM planning, we identifed two common failure modes (see also Figure \ref{fig:errors}):\\
\noindent \textbf{Subgoal mischaracterization:} Such errors arise when a subgoal isn't properly specified by the planner. That is, the plan subgoal is not a subset of the GTSG. When executing the step, this can lead to the resulting state not fulfilling the GTSG, and not affording the remainder of the plan. In practice, such errors arise from a lack of precision or missing information in the plan step, and are difficult to recover from as the plan step appears to have been executed correctly. To avoid such errors, we propose to introduce an \textit{expected outcomes} module, explicitly characterizing the plan subgoal and ensuring that it is a subset of the GTSG.\\
\noindent \textbf{Missing subgoal:} Missing subgoals arise from discrepancies between the semantics at the plan level and at the ground-truth level. This leads to plans that are correct, but cannot be achieved by simply mapping primitives to each plan step. This results in the execution module's output missing some primitives (for example not moving before trying to grasp an object), leading to options not being afforded. Such errors can be recovered from by implementing feedback mechanisms in the system, prompting the execution module to update the code by adding the missing primitives, compensating for the missing plan step. See Section \ref{sec:archi} for more details on these mechanisms.

\subsection{Architecture}\label{sec:archi}
We introduce the main components of our architecture. For each component, we describe the key idea, inputs and outputs. An illustration of the complete architecture can be found in Figure \ref{fig:top_fig} and prompts for each module can be found in the appendix.\\
\noindent\textbf{Planner:} The planning module receives the task instruction, expressed in language, and outputs a series of simple steps, close to the primitive level, achieving the task. In other words, the planner finds a set of subgoals for the agent to follow. In practice, we find that forming a plan before executing the task increases performance, facilitates recovery from errors and makes LLM-based task execution more explainable. It also reduces the likelihood of errors, as outputting code for the whole tasks typically has a higher chance of failure. While the idea of sequentially approaching a task is inspired by the chain-of-thought method\cite{cot}, we note that in our case the model responsible for planning is used in a zero-shot manner, instead of requiring examples of full reasoning chains.\\
\noindent\textbf{Expected outcomes module:} This module takes as input the instruction and the plan and outputs the expected outcome of each step of the plan, in terms of robot and world state. This is to provide a more precise characterization of each subgoal, resulting in a more complete plan. We also find that this module also improves the code quality of downstream modules, reducing hallucinations in code output.\\
\noindent\textbf{Execution module:} This module is in charge of executing the plan, step by step, using the motion primitives available to the agent. Its main inputs are the instruction, plan and expected outcomes. At each time step, the module outputs a python code snippet $C$ containing the code required to carry out the current step of the plan. In the module's first prompt, we also provide the headers and a use example of each motion primitive, and describe the setting of the experiments ('you are in charge of a robot with one arm'). At each step, we add to the context a follow-up prompt containing the updated plan and some feedback (see next paragraph).\\
\noindent\textbf{Feedback mechanism:} We implement a symbolic simulation with simple logic rules (for example, a robot cannot grasp an object if there is already an object in the gripper or if it is not at the same location as the object). Note that this simulation can be linked to the real world using perception modules, allowing for the feedback mechanism to be grounded. If the primitives execute successfully in the digital twin, they are executed in the environment. If they fail, appropriate feedback is returned to the execution module. At each time step, we also provide a reminder of the robot's location and of the object in the gripper. The goal of this mechanism is to provide the possibility to recover from code writing mistakes, primitive failure or planning issues by re-attempting a failed step. In Sections \ref{sec:exp} and \ref{sec:results}, we show that a few (two or three per skill) simple checks drastically robustify task execution.\\

\noindent \textbf{Implementation details:} We implement our architecture using ROS. The planner and expected outcomes module are in the same program, with both prompting separate instances of the LLM in turn when the task is published. When ready, the plan and expected outcomes are published by this program. The second program contains the execution module. When both the plan and expected outcomes have been received, the execution module receives its initial prompt. This prompt contains the task, plan, skills, a description of the setting (robot, number of arms) and environment (locations, objects). When appropriate (in our ablation study, we benchmark our full architecture with variants where the EO module is inactive, see Section \ref{sec:ablation}), the expected outcomes are also added to this prompt. When the execution module outputs some code, the code is published and executed in the digital twin, and the execution module receives feedback: either 'Done' if the code runs correctly or simple feedback when errors occur. The feedback also includes the robot's position and the object in the gripper. In the experiments where the feedback pipeline isn't active, the feedback is always 'Done'. If the step was successful, the current step is removed from the plan and the code is executed in the environment. Finally, the LLM is reprompted (by adding to the context), with the feedback, the plan and, if applicable, the current expected outcome, closing the loop.

\section{Experiments}\label{sec:exp}
This section describes our experimental protocol, in two robot settings: service and industrial.
\subsection{Service setting}
We consider a one-armed mobile manipulator robot helping a user in a home. The robot is used to carry out pick-and-place actions as might be required of a robotic home assistant. We randomly generate a set of $50$ pick-and-place tasks, each involving a random item, original item location and target item location. For each task, we programatically generate the instruction, randomly sampling among $3$ templates, and we use GPT-4o to generate two alternate task instructions. We instruct GPT to introduce variety in the statements, while ensuring that both the task and the information about the initial position of the objects are included. $10$ randomly selected instructions can be found in the appendix. We run an ablation study in order to verify the usefulness of our modules, as detailed in Section \ref{sec:ablation}.\\
The experiments are run using a symbolic simulator inspired by VirtualHome \cite{virtualhome, virtualhome2}, and we use VirtualHome to render videos of our experiments (see snapshots in Figure \ref{fig:service_exec}). In VirtualHome, the robot is represented by a humanoid agent. We implement a middleware between our architecture and the simulator, where the motion primitives and corresponding feedback mechanisms are implemented. We add a $10\%$ chance of failure in the grasp primitive. Experiments are run until the task is completed or the amount of interactions between the execution module and the environment becomes double the amount of steps in the plan (in which case the run has timed out and the task is failed).

\subsection{Industrial setting}
We consider a robot arm carrying out manipulation tasks inspired by the robothon task board \cite{taskboard}. This includes $2$ manipulation tasks and the full task board solution (press the blue button, plug in the probe cable, press the red button and open the trapdoor). The experiments in this setting were validated on a Franka arm, see Figure \ref{fig:charger_success} and the attached video (available on our git).\\
\noindent\textbf{Manipulation tasks:} We consider the following manipulation tasks
\begin{itemize}
    \item Plug the charger cable in the outlet.
    \item Put the voltage probe in its rack.
\end{itemize}
These tasks can only be carried out if the object is properly grasped before attempting to fulfil the task. For the charger task, the charger must be grasped by the plug part, while for the probe task, the probe must be grasped by the handle. We allow the agent to specify the part of the object on which the grasp is carried out by adding a parameter to the grasp primitive. We attempt the manipulation tasks using the \textbf{Full}, \textbf{FB}, \textbf{EO} and \textbf{Plan} versions of our architecture for these tasks.\\
\noindent\textbf{Task board:} We ask ten participants to ask the robot to solve the task board. We describe the task requirements but provide no details on the robot's skills. For these experiments, the primitives available to the robot represent the manipulation skills required by the task board: press a button, plug the probe cable and open the trapdoor, where the cable plugging skill was obtained from the manipulation task above, and we use our \textbf{full} architecture.

\subsection{Ablation study modalities}\label{sec:ablation}
We verify the effectiveness of our architecture with an ablation study. We consider the following architectures:
\begin{itemize}
\item \textbf{CaP:} Inspired by Code-as-Policies\cite{cap}, the LLM receives the instruction and the primitives and must output the full sequence of skills. We use GPT-4o for this approach.
\item \textbf{Plan:} There is no expected outcomes module, and the execution module's output is directly executed on the robot, without verification or feedback. We use LLaMA3.1 for the planner and execution module.
\item Expected outcomes \textbf{(EO):} As above, there is no feedback. The expected outcomes module is active and the expected outcomes are given as input to the execution module. All modules use LLaMA3.1.
\item \textbf{FB:} The execution module receives feedback, but the expected outcomes module is not active. Both modules use LLaMA3.1.
\item \textbf{Full:} This is the full architecture as described in Section \ref{sec:archi}. All modules use LLaMA3.1.
\item \textbf{DS full}: Additionally, we implement a variation of the \textbf{Full} architecture, where we use the Deepseek-r1:7b model instead of LLama. Aside from a few modifications to the parsing mechanisms to accommodate Deepseek's output format, the code is the same as in the \textbf{Full} experiments.
\end{itemize}
\section{Results and discussion}\label{sec:results}
In this section, we show that our architecture significantly outperforms ablated versions, even surpassing the GPT-based code-as-policies baseline. We also demonstrate its suitability to real robot use through a hardware experiment. By studying the amount of compute (average tokens and execution module inferences) required for the service tasks, we show how our modules result in faster interaction, facilitating data generation.

\subsection{Service robot results}
\begin{figure}
    \centering
    \vspace{0.2cm}
    \includegraphics[width=0.4\columnwidth]{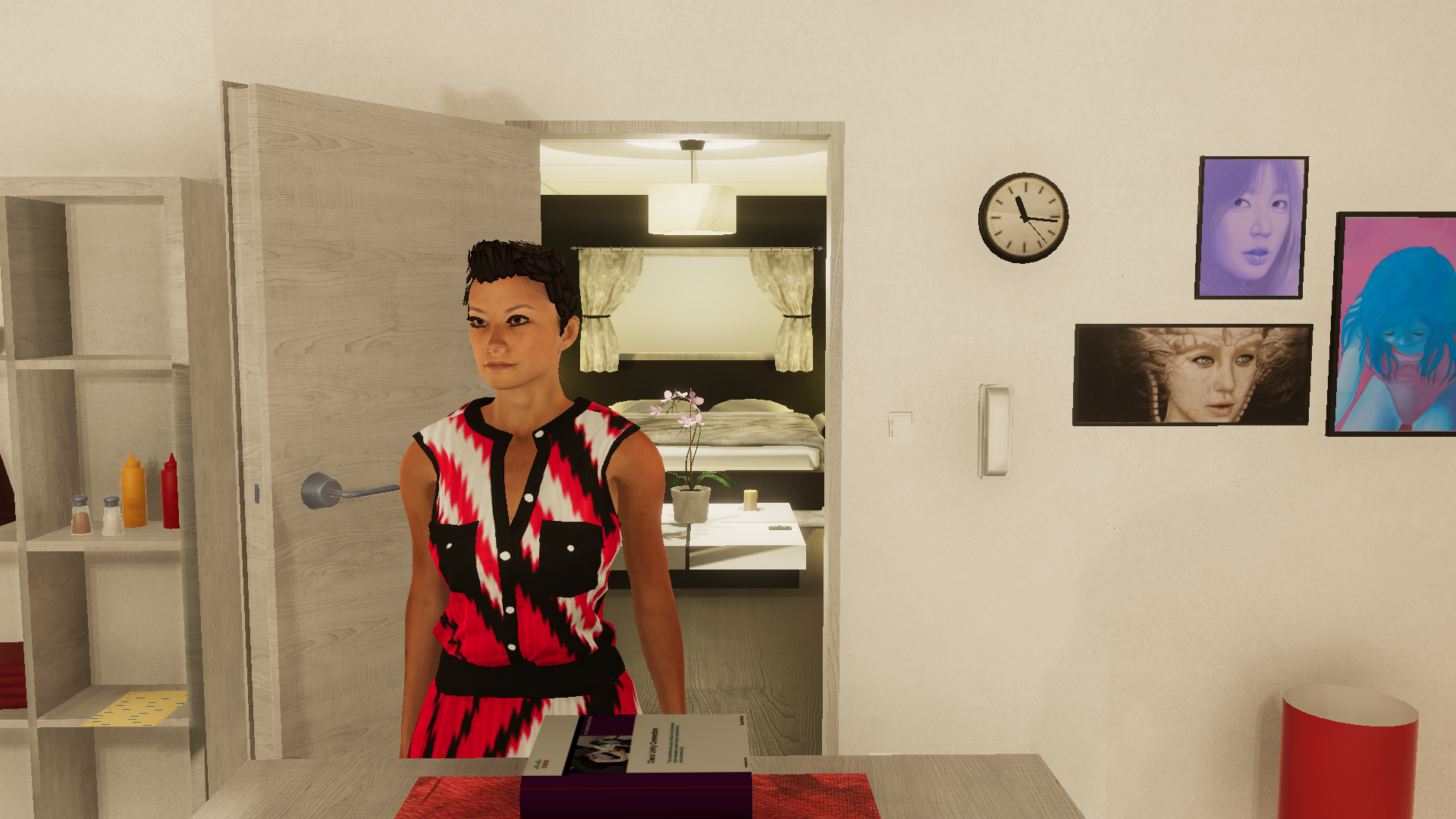}\hspace{0.1cm}\includegraphics[width=0.4\columnwidth]{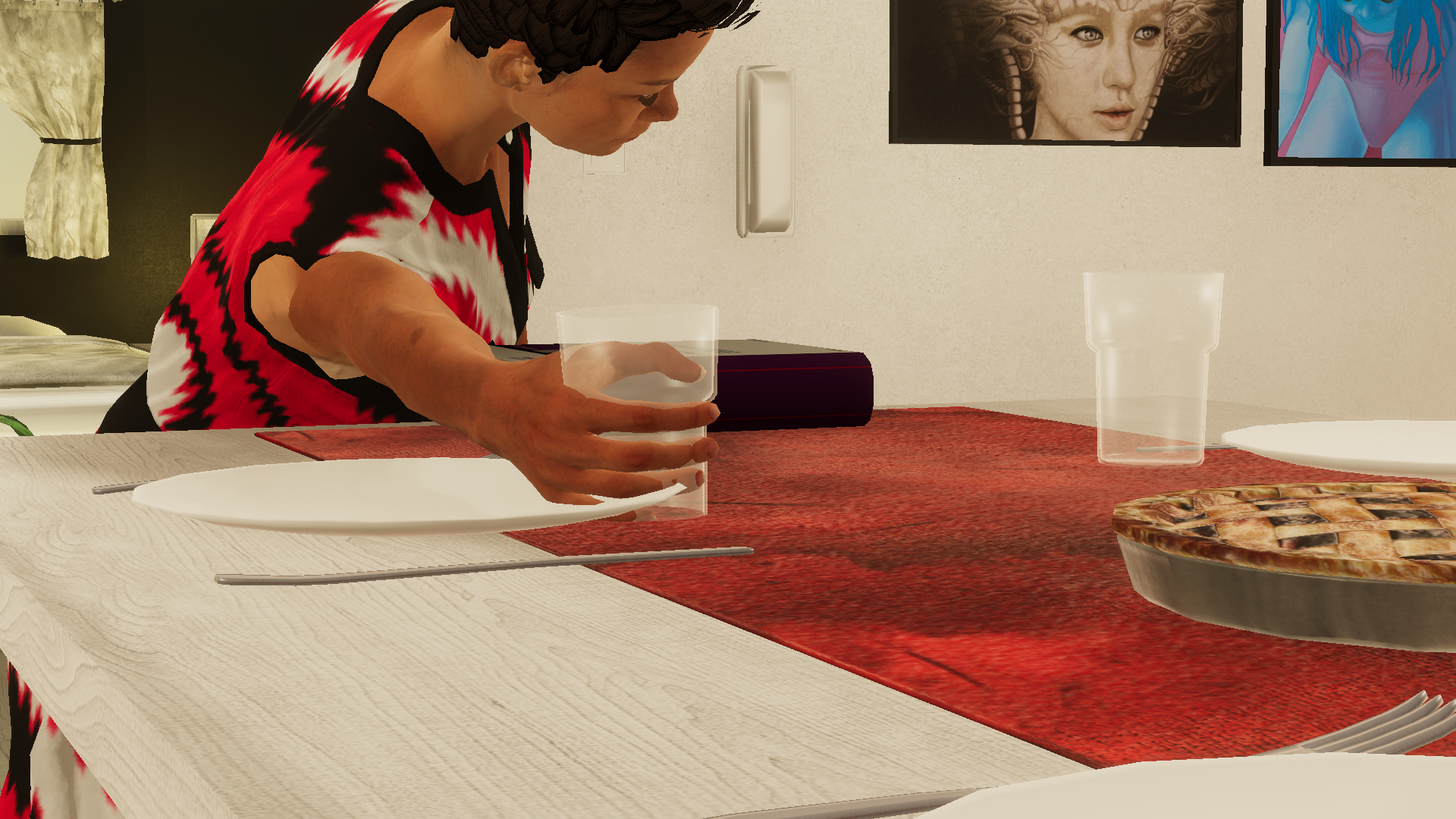}\vspace{0.05cm}\\
    \includegraphics[width=0.4\columnwidth]{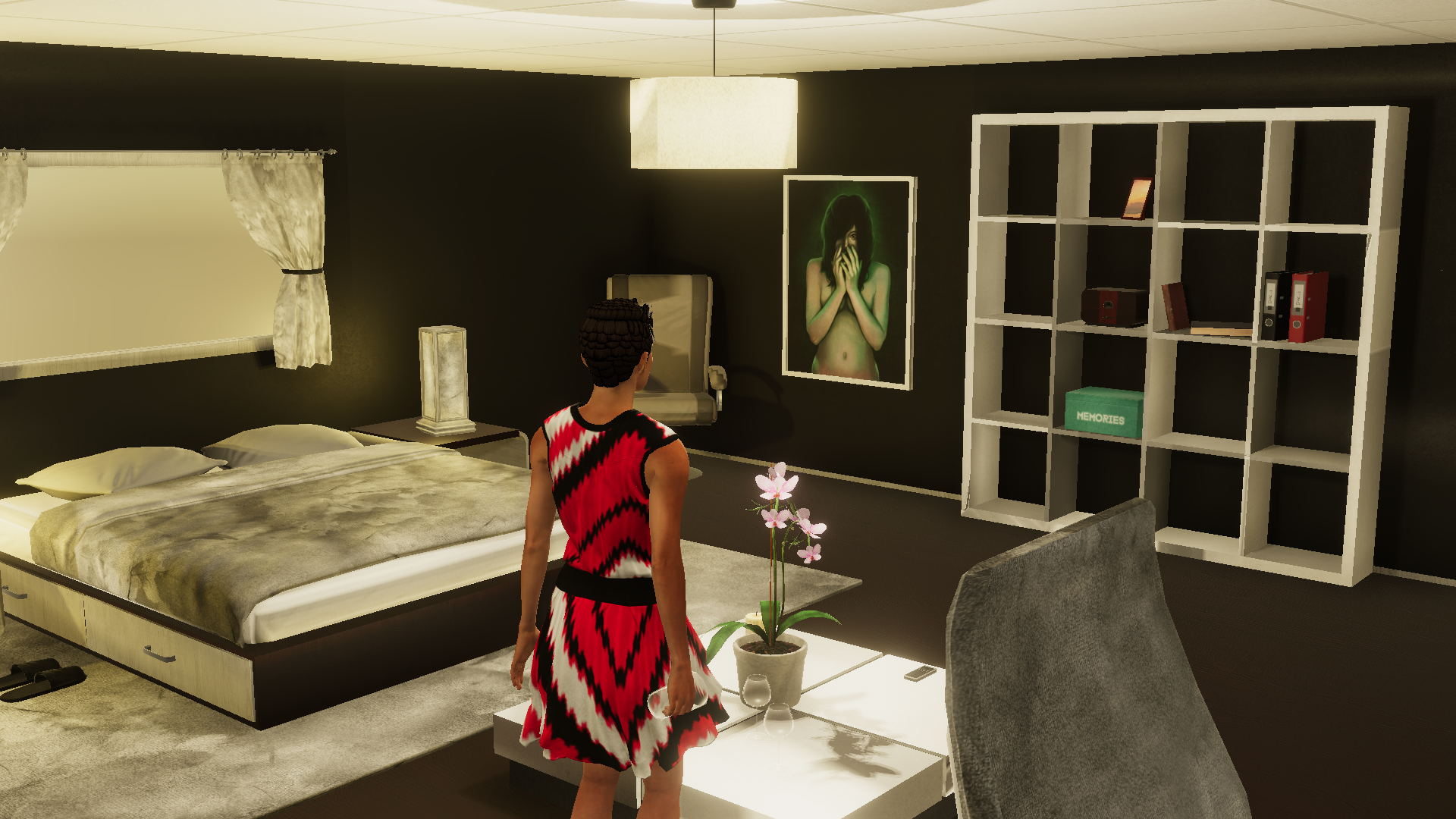}\hspace{0.1cm}\includegraphics[width=0.4\columnwidth]{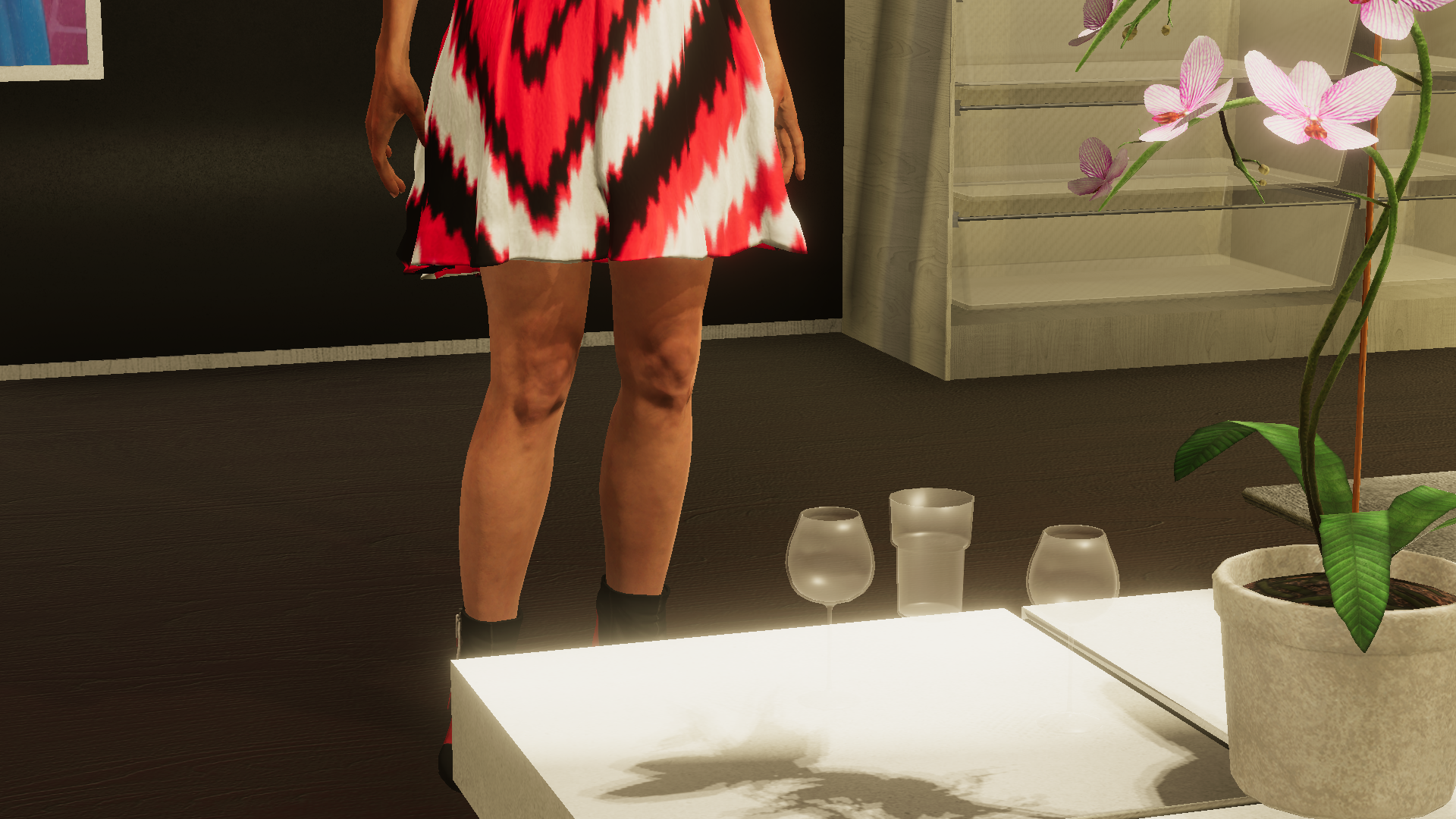}
    \caption{Successful execution of the "Move the Water Glass to the Coffee table. It is currently on the Kitchen table." task. From left to right, top to bottom: (i) the service robot approaches the kitchen table; (ii) the robot grasps the glass of water; (iii) the robot moves to the coffee table, (iv) the task ends with the robot putting down the glass.}
    \label{fig:service_exec}
\end{figure}

We run the experiment corpus three times, using the different instruction sets. A mean, std summary of the success rates of each architecture variant can be found in Table \ref{tab:success_rates}.\\
Over all tasks, \textbf{our full architecture strongly outperforms its ablated versions}. Individually, the components of our architecture each improve the success rate over the \textbf{plan} version. Interestingly, we notice that by precisely characterizing each subgoal, the expected outcomes module increases the code quality of the execution module's output, leading to a sharp increase in success rate. Simulation snapshots taken from a successful execution of the task "Move the Water Glass to the Coffee table. It is currently on the Kitchen table.", using our full architecture, are shown in Figure \ref{fig:service_exec}.\\
Despite GPT-4o' typical performance at code generation tasks, \textbf{the code-as-policy approach obtains a lower success rate than our architecture}. The two main causes for this are a lack of grounding between the LLM and the environment and the open-loop execution failing to recover from grasp failure. We also note that the Deepseek-based architecture almost never succeeds. We conjecture that this is due to the model being specialized in one-shot reasoning and problem solving, making it less suitable for long-horizon interactions.

\begin{table}[]
    \centering
    \begin{tabular}{|l|c|c|c|}
    \hline
        Architecture & Success rate & Execution calls & Tokens output \\
        \hline
        CaP & 92.67 $\pm$ 3.40 & 1 $\pm$ 0 & 88.13 $\pm$ 0.98\\
        Plan & 10.67 $\pm$ 3.40 & 4.11 $\pm$ 0.02 & 386.67 $\pm$ 15.11\\
        EO & 87.33 $\pm$ 0.94 & 4.11 $\pm$ 0.02 & 283.20 $\pm$ 11.46\\
        FB & 67.33 $\pm$ 6.18 & 6.11 $\pm$ 0.12 & 547.40 $\pm$ 73.71\\
        Full & 96.00 $\pm$ 2.83 & 5.19 $\pm$ 0.47 & 320.00 $\pm$ 30.80\\
        Full DS & 17.33 $\pm$ 0.94 & 6.96 $\pm$ 0.22 & 3627.67 $\pm$ 241.17\\
        \hline
    \end{tabular}
    \caption{Percentage success rate (mean $\pm$ std), average number of calls to the execution module and average number of tokens output by the execution module for each of the architectures in our benchmark. Our full architecture outperforms ablated versions, and surpasses the GPT-based code-as-policies method. Both the expected outcomes module and feedback mechanism increase success rate, and the expected outcomes module reduces the amount of execution module calls and the number of tokens output by the model, resulting in faster execution.}
    \label{tab:success_rates}
\end{table}

\subsection{Industrial robot results}
\begin{figure}
    \centering
    \vspace{0.2cm}
    \includegraphics[width=0.4\columnwidth]{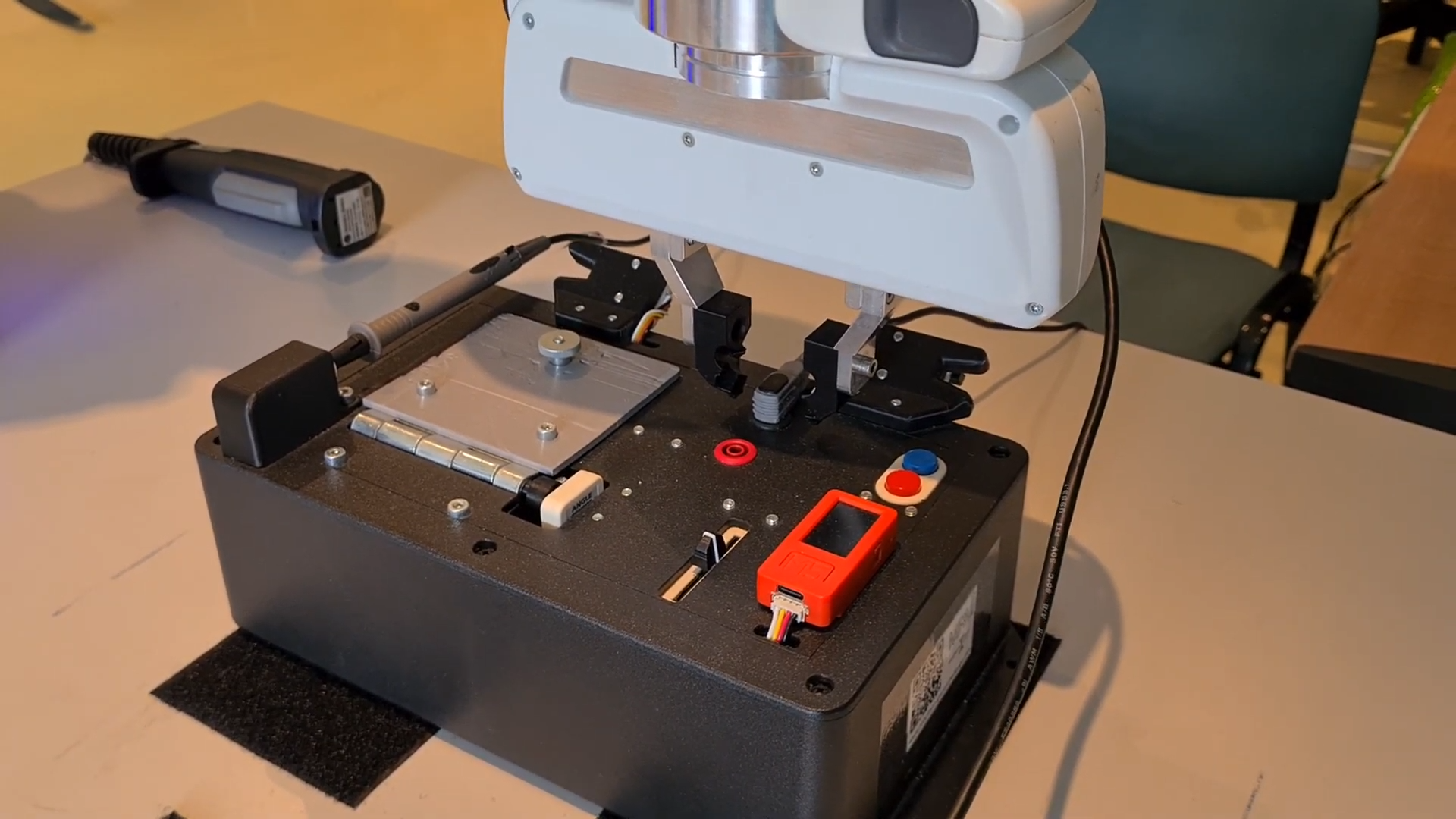}\hspace{0.1cm}\includegraphics[width=0.4\columnwidth]{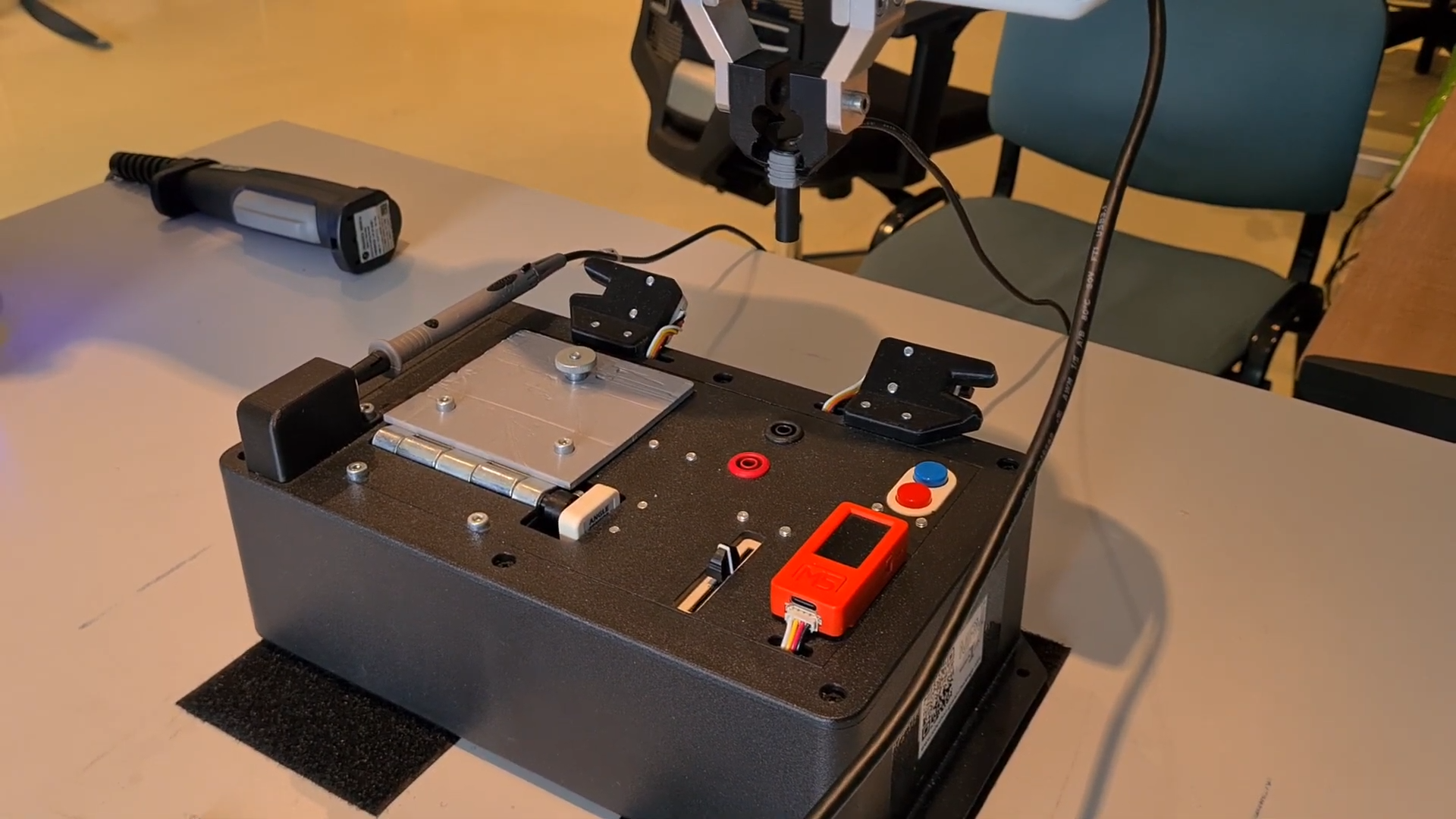}\vspace{0.1cm}\\
    \includegraphics[width=0.4\columnwidth]{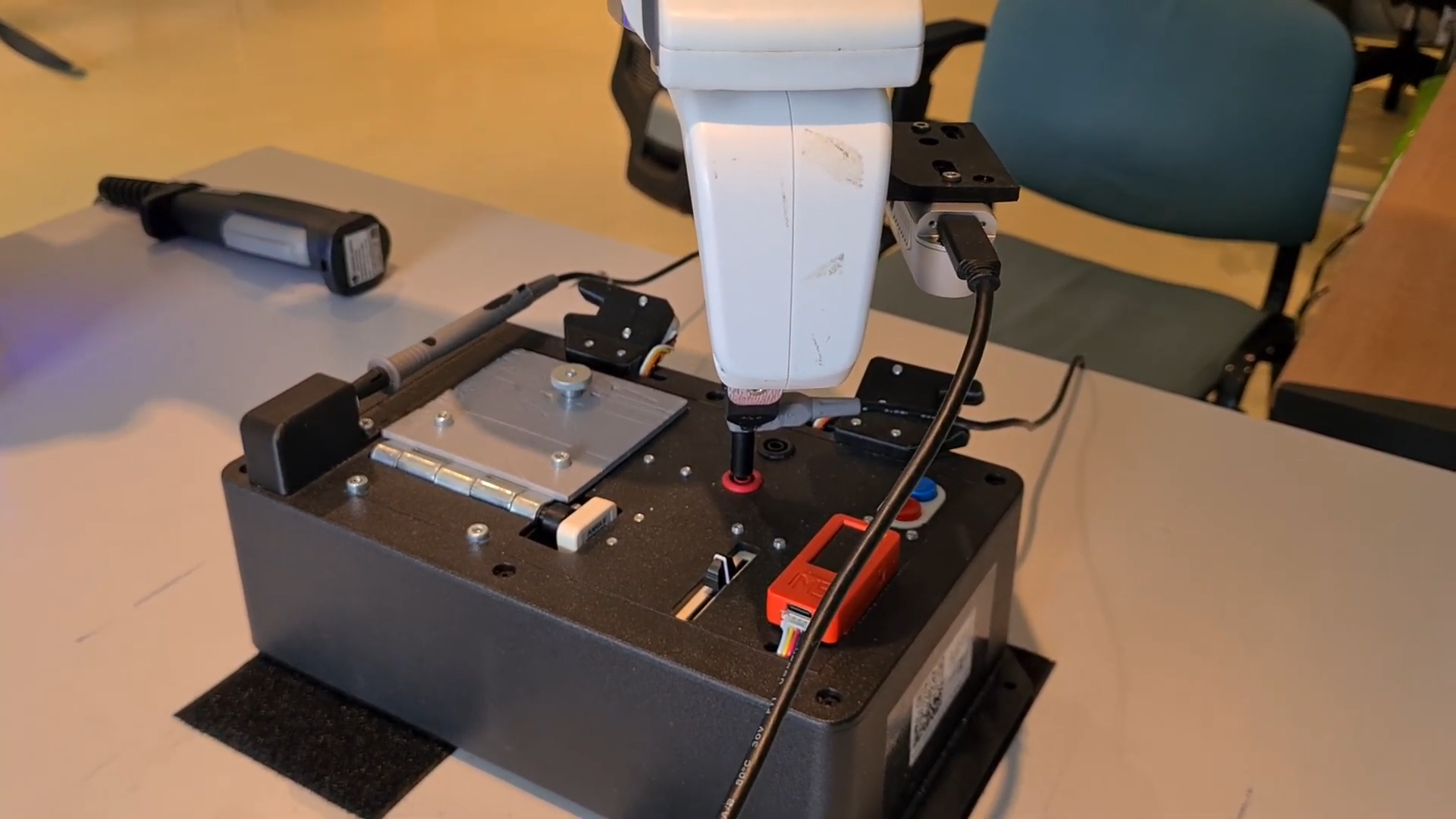}\hspace{0.1cm}\includegraphics[width=0.4\columnwidth]{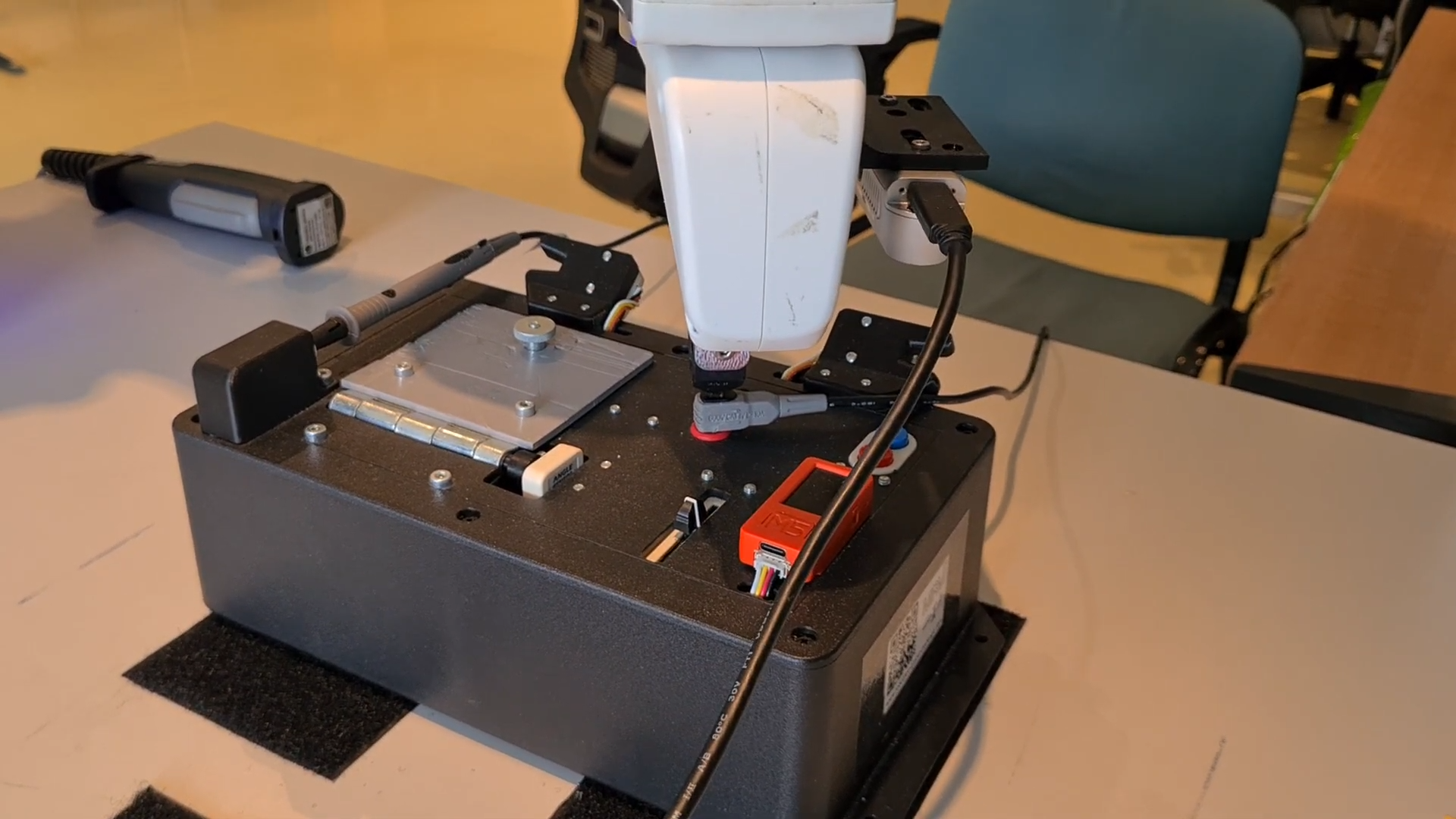}
    \caption{Successful execution of the "Plug in the charger" task. From left to right, top to bottom: (i) the gripper approaches the charger's plug; (ii) the charger is grasped by the gripper; (iii) the arm begins plugging the charger, (iv) the task ends with the charger plugged.}
    \label{fig:charger_success}
\end{figure}

Our full pipeline succeeds in \textbf{solving the task board} with each of the $10$ participants instructions, and in \textbf{carrying out both manipulation tasks}. When using the \textbf{Plan} and \textbf{FB} architectures, the success rate in the manipulation tasks drops to $0.5$, due to the execution module failing to properly specify which part of the object should be grasped. This highlights how the expected outcomes module can be used to improve subgoal characterization (in this case by specifying grasp targets on the objects), leading to more robust execution.\\
These hardware experiments (see Figure \ref{fig:charger_success} and the video attachment to this paper) show that our architecture is suitable for real-time use on a real robot.

\subsection{Computation and inference time}
In our service experiments, when using our full architecture, our execution module outputs on average $3.64$ times more tokens than the GPT-based architecture (see Table \ref{tab:success_rates}). It is estimated GPT-4o is a mixture of $110$ billion parameter experts (MoE), as opposed to the $8$ billion of LLama3.1, which is $13.75$ times smaller. As the amount of floating-point operations is approximately linear with the model (or expert) size and the number of output tokens, it results that our method needs about \textbf{3.78 times less compute} operations per task compared to a GPT-based architecture. Further, our execution module inference time is on average \textbf{2.05 seconds} on a standard ($8$GB VRAM) GPU, showing the suitability of our approach for generating large datasets and for online robotic interaction.

\section{Concluding remarks and future work}
In this paper, we formalized the problem of planning and executing robotic tasks with large language models as the problem of following a set of subgoals through a chain of options in a goal-conditioned POMDP. We showed that execution errors of such a pipeline can be classified within this formalism as missing or mischaracterized subgoals preventing option chaining. Then, we proposed an architecture with technical components designed to prevent such errors. Namely, we added a second planning component designed to explicitly characterize the outcome of each plan step and implemented some feedback mechanisms. Through experiments in a service robot setting, we showed that our architecture achieves robust execution, while incurring low computational costs (3.78 times lower than a GPT-based baseline), for pick-and-place tasks, and experiments in an industrial robot setting show that our pipeline also succeeds at manipulation tasks. Our hardware validation shows that our architecture can be applied in real time using a standard computer and an openly accessible language model, providing encouraging insights into the applicability of smaller language models for robotics.\\
Future research directions include testing our architecture as part of a robotic cognitive architecture and fine-tuning models for robotic planning with data obtained using our pipeline.

\section*{Acknowledgements}
This work was supported by the European Union Horizon Europe Framework Programme, through the PILLAR (grant agreement 101070381) and euROBIN (grant agreement 101070596) projects.

\bibliographystyle{IEEEtran}
\bibliography{sample}

\begin{thebibliography}{10}
\providecommand{\url}[1]{#1}
\csname url@samestyle\endcsname
\providecommand{\newblock}{\relax}
\providecommand{\bibinfo}[2]{#2}
\providecommand{\BIBentrySTDinterwordspacing}{\spaceskip=0pt\relax}
\providecommand{\BIBentryALTinterwordstretchfactor}{4}
\providecommand{\BIBentryALTinterwordspacing}{\spaceskip=\fontdimen2\font plus
\BIBentryALTinterwordstretchfactor\fontdimen3\font minus \fontdimen4\font\relax}
\providecommand{\BIBforeignlanguage}[2]{{%
\expandafter\ifx\csname l@#1\endcsname\relax
\typeout{** WARNING: IEEEtran.bst: No hyphenation pattern has been}%
\typeout{** loaded for the language `#1'. Using the pattern for}%
\typeout{** the default language instead.}%
\else
\language=\csname l@#1\endcsname
\fi
#2}}
\providecommand{\BIBdecl}{\relax}
\BIBdecl

\bibitem{you2023robot}
H.~You, Y.~Ye, T.~Zhou, Q.~Zhu, and J.~Du, ``Robot-enabled construction assembly with automated sequence planning based on {ChatGPT: RoboGPT},'' \emph{Buildings}, vol.~13, no.~7, p. 1772, 2023.

\bibitem{llmpp}
B.~Liu, Y.~Jiang, X.~Zhang, Q.~Liu, S.~Zhang, J.~Biswas, and P.~Stone, ``Llm+ p: Empowering large language models with optimal planning proficiency,'' \emph{arXiv preprint arXiv:2304.11477}, 2023.

\bibitem{xie2023translating}
Y.~Xie, C.~Yu, T.~Zhu, J.~Bai, Z.~Gong, and H.~Soh, ``Translating natural language to planning goals with large-language models,'' \emph{arXiv preprint arXiv:2302.05128}, 2023.

\bibitem{pddlllm}
T.~Silver, V.~Hariprasad, R.~S. Shuttleworth, N.~Kumar, T.~Lozano-P{\'e}rez, and L.~P. Kaelbling, ``{PDDL} planning with pretrained large language models,'' in \emph{NeurIPS 2022 foundation models for decision making workshop}, 2022.

\bibitem{LLMbt}
H.~Zhou, Y.~Lin, L.~Yan, J.~Zhu, and H.~Min, ``{LLM-BT}: Performing robotic adaptive tasks based on large language models and behavior trees,'' \emph{arXiv preprint arXiv:2404.05134}, 2024.

\bibitem{sayplan}
K.~Rana, J.~Haviland, S.~Garg, J.~Abou-Chakra, I.~Reid, and N.~Suenderhauf, ``Sayplan: Grounding large language models using 3d scene graphs for scalable task planning,'' \emph{arXiv preprint arXiv:2307.06135}, 2023.

\bibitem{cap}
J.~Liang, W.~Huang, F.~Xia, P.~Xu, K.~Hausman, B.~Ichter, P.~Florence, and A.~Zeng, ``Code as policies: Language model programs for embodied control,'' in \emph{2023 IEEE International Conference on Robotics and Automation (ICRA)}.\hskip 1em plus 0.5em minus 0.4em\relax IEEE, 2023, pp. 9493--9500.

\bibitem{human_aware}
Y.~Liu, L.~Palmieri, S.~Koch, I.~Georgievski, and M.~Aiello, ``Towards human awareness in robot task planning with large language models,'' \emph{arXiv preprint arXiv:2404.11267}, 2024.

\bibitem{progprompt}
I.~Singh, V.~Blukis, A.~Mousavian, A.~Goyal, D.~Xu, J.~Tremblay, D.~Fox, J.~Thomason, and A.~Garg, ``Progprompt: Generating situated robot task plans using large language models,'' in \emph{2023 IEEE International Conference on Robotics and Automation (ICRA)}.\hskip 1em plus 0.5em minus 0.4em\relax IEEE, 2023, pp. 11\,523--11\,530.

\bibitem{sup_ddown}
H.~Ha, P.~Florence, and S.~Song, ``Scaling up and distilling down: Language-guided robot skill acquisition,'' in \emph{Conference on Robot Learning}.\hskip 1em plus 0.5em minus 0.4em\relax PMLR, 2023, pp. 3766--3777.

\bibitem{planning_then_code}
Y.~Ouyang, J.~Li, Y.~Li, Z.~Li, C.~Yu, K.~Sreenath, and Y.~Wu, ``Long-horizon locomotion and manipulation on a quadrupedal robot with large language models,'' \emph{arXiv preprint arXiv:2404.05291}, 2024.

\bibitem{robohorizon}
Z.~Chen, J.~Huo, Y.~Chen, and Y.~Gao, ``Robohorizon: An llm-assisted multi-view world model for long-horizon robotic manipulation,'' \emph{arXiv preprint arXiv:2501.06605}, 2025.

\bibitem{copa}
H.~Huang, F.~Lin, Y.~Hu, S.~Wang, and Y.~Gao, ``{CoPa}: General robotic manipulation through spatial constraints of parts with foundation models,'' in \emph{2024 IEEE/RSJ International Conference on Intelligent Robots and Systems (IROS)}, 2024, pp. 9488--9495.

\bibitem{autort}
M.~Ahn, D.~Dwibedi, C.~Finn, M.~G. Arenas, K.~Gopalakrishnan, K.~Hausman, B.~Ichter, A.~Irpan, N.~Joshi, R.~Julian \emph{et~al.}, ``{AutoRT}: Embodied foundation models for large scale orchestration of robotic agents,'' \emph{arXiv preprint arXiv:2401.12963}, 2024.

\bibitem{octo}
O.~M. Team, D.~Ghosh, H.~Walke, K.~Pertsch, K.~Black, O.~Mees, S.~Dasari, J.~Hejna, T.~Kreiman, C.~Xu \emph{et~al.}, ``Octo: An open-source generalist robot policy,'' \emph{arXiv preprint arXiv:2405.12213}, 2024.

\bibitem{openvla}
\BIBentryALTinterwordspacing
M.~J. Kim, K.~Pertsch, S.~Karamcheti, T.~Xiao, A.~Balakrishna, S.~Nair, R.~Rafailov, E.~Foster, G.~Lam, P.~Sanketi, Q.~Vuong, T.~Kollar, B.~Burchfiel, R.~Tedrake, D.~Sadigh, S.~Levine, P.~Liang, and C.~Finn, ``{OpenVLA}: An open-source vision-language-action model,'' 2024. [Online]. Available: \url{https://arxiv.org/abs/2406.09246}
\BIBentrySTDinterwordspacing

\bibitem{chatvla}
Z.~Zhou, Y.~Zhu, M.~Zhu, J.~Wen, N.~Liu, Z.~Xu, W.~Meng, R.~Cheng, Y.~Peng, C.~Shen \emph{et~al.}, ``Chat{VLA}: Unified multimodal understanding and robot control with vision-language-action model,'' \emph{arXiv preprint arXiv:2502.14420}, 2025.

\bibitem{pi_0}
K.~Black, N.~Brown, D.~Driess, A.~Esmail, M.~Equi, C.~Finn, N.~Fusai, L.~Groom, K.~Hausman, B.~Ichter \emph{et~al.}, ``$\pi_0 $: A vision-language-action flow model for general robot control,'' \emph{arXiv preprint arXiv:2410.24164}, 2024.

\bibitem{brohan2023can}
A.~Brohan, Y.~Chebotar, C.~Finn, K.~Hausman, A.~Herzog, D.~Ho, J.~Ibarz, A.~Irpan, E.~Jang, R.~Julian \emph{et~al.}, ``Do as i can, not as i say: Grounding language in robotic affordances,'' in \emph{Conference on robot learning}.\hskip 1em plus 0.5em minus 0.4em\relax PMLR, 2023, pp. 287--318.

\bibitem{rana2023sayplan}
K.~Rana, J.~Haviland, S.~Garg, J.~Abou-Chakra, I.~Reid, and N.~Suenderhauf, ``Sayplan: Grounding large language models using 3d scene graphs for scalable task planning,'' \emph{arXiv preprint arXiv:2307.06135}, 2023.

\bibitem{rajvanshi2024saynav}
A.~Rajvanshi, K.~Sikka, X.~Lin, B.~Lee, H.-P. Chiu, and A.~Velasquez, ``Saynav: Grounding large language models for dynamic planning to navigation in new environments,'' in \emph{Proceedings of the International Conference on Automated Planning and Scheduling}, vol.~34, 2024, pp. 464--474.

\bibitem{tong2024oval}
E.~Tong, A.~Opipari, S.~Lewis, Z.~Zeng, and O.~C. Jenkins, ``Oval-prompt: Open-vocabulary affordance localization for robot manipulation through llm affordance-grounding,'' \emph{arXiv preprint arXiv:2404.11000}, 2024.

\bibitem{closed_loop}
P.~Zhi, Z.~Zhang, M.~Han, Z.~Zhang, Z.~Li, Z.~Jiao, B.~Jia, and S.~Huang, ``Closed-loop open-vocabulary mobile manipulation with gpt-4v,'' \emph{arXiv preprint arXiv:2404.10220}, 2024.

\bibitem{ren2023robots}
A.~Z. Ren, A.~Dixit, A.~Bodrova, S.~Singh, S.~Tu, N.~Brown, P.~Xu, L.~Takayama, F.~Xia, J.~Varley \emph{et~al.}, ``Robots that ask for help: Uncertainty alignment for large language model planners,'' \emph{arXiv preprint arXiv:2307.01928}, 2023.

\bibitem{mullen2024towards}
J.~F. Mullen~Jr and D.~Manocha, ``Towards robots that know when they need help: Affordance-based uncertainty for large language model planners,'' \emph{arXiv preprint arXiv:2403.13198}, 2024.

\bibitem{inner_monologue}
W.~Huang, F.~Xia, T.~Xiao, H.~Chan, J.~Liang, P.~Florence, A.~Zeng, J.~Tompson, I.~Mordatch, Y.~Chebotar \emph{et~al.}, ``Inner monologue: Embodied reasoning through planning with language models,'' \emph{arXiv preprint arXiv:2207.05608}, 2022.

\bibitem{RoboGPT}
Y.~Chen, W.~Cui, Y.~Chen, M.~Tan, X.~Zhang, J.~Liu, H.~Li, D.~Zhao, and H.~Wang, ``{RoboGPT}: an llm-based long-term decision-making embodied agent for instruction following tasks,'' \emph{IEEE Transactions on Cognitive and Developmental Systems}, pp. 1--11, 2025.

\bibitem{glam}
\BIBentryALTinterwordspacing
T.~Carta, C.~Romac, T.~Wolf, S.~Lamprier, O.~Sigaud, and P.-Y. Oudeyer, ``Grounding large language models in interactive environments with online reinforcement learning,'' 2024. [Online]. Available: \url{https://arxiv.org/abs/2302.02662}
\BIBentrySTDinterwordspacing

\bibitem{malmm}
H.~Singh, R.~J. Das, M.~Han, P.~Nakov, and I.~Laptev, ``{MALMM}: Multi-agent large language models for zero-shot robotics manipulation,'' \emph{arXiv preprint arXiv:2411.17636}, 2024.

\bibitem{hicrisp}
C.~Ming, J.~Lin, P.~Fong, H.~Wang, X.~Duan, and J.~He, ``{HiCRISP}: An llm-based hierarchical closed-loop robotic intelligent self-correction planner,'' in \emph{2024 China Automation Congress (CAC)}.\hskip 1em plus 0.5em minus 0.4em\relax IEEE, 2024, pp. 4310--4315.

\bibitem{gcmdp}
T.~Schaul, D.~Horgan, K.~Gregor, and D.~Silver, ``Universal value function approximators,'' in \emph{International conference on machine learning}.\hskip 1em plus 0.5em minus 0.4em\relax PMLR, 2015, pp. 1312--1320.

\bibitem{sutton_options}
R.~S. Sutton, D.~Precup, and S.~Singh, ``Between {MDPs} and semi-{MDPs}: A framework for temporal abstraction in reinforcement learning,'' \emph{Artificial intelligence}, vol. 112, no. 1-2, pp. 181--211, 1999.

\bibitem{affordance_review}
P.~Ard{\'o}n, E.~Pairet, K.~S. Lohan, S.~Ramamoorthy, and R.~Petrick, ``Building affordance relations for robotic agents-a review,'' \emph{arXiv preprint arXiv:2105.06706}, 2021.

\bibitem{deep_affordance_RL}
X.~Yang, Z.~Ji, J.~Wu, and Y.-K. Lai, ``Recent advances of deep robotic affordance learning: a reinforcement learning perspective,'' \emph{IEEE Transactions on Cognitive and Developmental Systems}, vol.~15, no.~3, pp. 1139--1149, 2023.

\bibitem{cot}
J.~Wei, X.~Wang, D.~Schuurmans, M.~Bosma, F.~Xia, E.~Chi, Q.~V. Le, D.~Zhou \emph{et~al.}, ``Chain-of-thought prompting elicits reasoning in large language models,'' \emph{Advances in neural information processing systems}, vol.~35, pp. 24\,824--24\,837, 2022.

\bibitem{virtualhome}
X.~Puig, K.~Ra, M.~Boben, J.~Li, T.~Wang, S.~Fidler, and A.~Torralba, ``Virtualhome: Simulating household activities via programs,'' in \emph{Proceedings of the IEEE Conference on Computer Vision and Pattern Recognition}, 2018, pp. 8494--8502.

\bibitem{virtualhome2}
\BIBentryALTinterwordspacing
X.~Puig, T.~Shu, S.~Li, Z.~Wang, Y.-H. Liao, J.~B. Tenenbaum, S.~Fidler, and A.~Torralba, ``Watch-and-help: A challenge for social perception and human-ai collaboration,'' 2021. [Online]. Available: \url{https://arxiv.org/abs/2010.09890}
\BIBentrySTDinterwordspacing

\bibitem{taskboard}
P.~So, A.~Sarabakha, F.~Wu, U.~Culha, F.~J. Abu-Dakka, and S.~Haddadin, ``Digital robot judge: Building a task-centric performance database of real-world manipulation with electronic task boards,'' \emph{IEEE Robotics \& Automation Magazine}, 2024.

\end{thebibliography}

%\appendix
\begin{appendices}

\section{Prompts}
We report the prompt templates used for our service experiments. The prompts used in the industrial setting are the same up to minor adaptations, and can be found on our github. The colored text segments are programatically added to the prompts:
\begin{itemize}
    \item \textcolor{purple}{Task}: Task instruction
    \item \textcolor{purple}{Plan}: Plan generated by the planning module, as a Python-style list of strings
    \item \textcolor{purple}{EOs}: Expected outcomes generated by the EO module, as a Python-style dictionary of strings
    \item \textcolor{purple}{Skills}: Python-style headers for the functions containing the motion primitives
    \item \textcolor{purple}{Environment}: Name of the objects and locations in the environment (in Python string format), and location of the objects and robot at the beginning of the trial.
\end{itemize}
\noindent \textbf{Planner module prompt:}\\
\texttt{\small{You are in charge of a mobile robot with an arm ending in a gripper. Your task is the following: \textcolor{purple}{Task}\\
Please output an plan, composed of simple actions, to carry out this task. Remember that the robot should always move to a location before interacting with objects in this location, unless it is already there. However, you can assume that simple actions (such as grasping or putting down objects) automatically move the arm to the correct position.\\
Please only output the plan as a tuple of strings, where each step is a string, without any other text.}}\\
\noindent \textbf{Expected outcomes module prompt:}\\
\texttt{\small{You are in charge of executing the following task: \textcolor{purple}{Task}. The plan consists of the following steps: \textcolor{purple}{Plan} Each of the steps of the plan will be executed with a mobile robot equipped with an arm ending in a gripper. For each step of the plan, I need you to give the expected outcome of the actions involved in the step, in physical and visual terms.\\
This should consist of one or two short, simple sentences that are a more complete and detailed description of the step's outcome. The sentences should describe the final state of the robot, for example if it should be at a location, have grasped an object (and what part of the object, if relevant for the task), or where an object should be put down. You can add some information if the plan is too concise. Here are some examples, with the plan step first and the expected outcome after:\\
- Put bottle on shelf: The bottle should be on the shelf.\\
- Grasp the mug: The mug should be in the robot's gripper.\\
- Grasp the knife: The knife blade should be in the robot's gripper.\\
For each step of the plan, please briefly describe the expected outcome as shown above. Please try to be concise and focus on the most relevant information. Please fill out the following python dictionary with the expected outcomes: \textcolor{purple}{dictionary template}. Only output the dictionary and no other text.}}\\
\noindent \textbf{Execution module prompt:}\\
\texttt{\small{Context:\\
Your are now in charge of a mobile robot equipped with one arm with a parallel gripper. You will be given a high-level task that you will need to fulfill using this robot, and the corresponding plan, which is a series of simpler steps. You will need to carry out the task step by step by interacting with the system using some code primitives. At each step the plan will be updated and you will receive feedback.
The skills are python functions, which allow you to perceive and act on your environment.\\
Skills:\\
Here are the functions and skills, with examples of the syntax:\\
\textcolor{purple}{skills}\\
The task and the plan:
You are in charge of executing the following task: \textcolor{purple}{Task}. The plan is: \textcolor{purple}{Plan}\\
Here are the expected outcomes of each step in the plan, which you can use as a guide:\\
\textcolor{purple}{EOs}\\
\textcolor{purple}{Environment} \\
What I need you to do:\\
Please define a function do(), which will contain mostly action primitives to solve the steps of the plan one by one. Please output python code, enclosed between the tags <code> and </code>. Please only use the functions I defined above and ensure the locations and objects that you pass as arguments are correct.}}

\section{Tasks}
Here are $10$ randomly-selected sample task instructions from the service robot corpus:
\begin{itemize}
    \item Move the Water Glass to the Coffee table. It is currently on the Kitchen table.
    \item Move the Pills from the Desk to the Kitchen counter
    \item Move the Fork to the Kitchen table. It is on the Coffee table.
    \item Put the Mouse on the Table. The Mouse is on the Desk.
    \item Transfer the Knife from the Desk to the Kitchen table.
    \item Set the Screwdriver down on the Coffee table. It is now on the Desk.
    \item Place the Plate onto the Kitchen counter. It is on the Coffee table.
    \item Retrieve the Knife from the Coffee table and place it on the Desk.
    \item Take the Cupcake and put it on the Desk. The Cupcake is on the Table.
    \item Move the Fork onto the Table. It is on the Desk.
\end{itemize}

\end{appendices}

\end{document}